\documentclass{article}

\usepackage{arxiv}

\usepackage[utf8]{inputenc} 
\usepackage[T1]{fontenc}    
\usepackage{hyperref}       
\usepackage{url}            
\usepackage{booktabs}       
\usepackage{amsfonts}       
\usepackage{nicefrac}       
\usepackage{microtype}      
\usepackage{lipsum}
\usepackage{graphicx}
\graphicspath{ {./images/} }
\usepackage{xcolor}         
\usepackage{natbib}

\definecolor{darkgreen}{rgb}{0.0, 0.5, 0.0}
\usepackage{algorithm}
\usepackage{algorithmic}
\usepackage{bm}
\usepackage{makecell}
\usepackage{amsthm}
\usepackage{amssymb}
\usepackage{amsmath}

\usepackage{wrapfig}
\usepackage{graphicx}
\usepackage{subfigure}
\usepackage{diagbox}
\usepackage{multirow} 

\theoremstyle{plain}
\newtheorem{theorem}{Theorem}

\theoremstyle{definition}

\usepackage{tcolorbox}
\theoremstyle{remark} 
\newtheorem{remark}[theorem]{Remark}

\usepackage{colortbl,color}
\definecolor{greyC}{RGB}{180,180,180}
\definecolor{greyL}{RGB}{235,235,235}

\definecolor{citeColor}{RGB}{0,20,115}
\hypersetup{colorlinks,linkcolor={red},citecolor={citeColor},urlcolor={citeColor}}

\title{Intra-Trajectory Consistency for Reward Modeling}

\author{%
  Chaoyang Zhou \\
  Wuhan University \\
  \texttt{zhoucy@whu.edu.cn} \\
  \And
  Shunyu Liu \\
  Nanyang Technological University \\
  \texttt{shunyu.liu@ntu.edu.sg} \\
  \And
  Zengmao Wang\thanks{Corresponding author.} \\
  Wuhan University \\
  \texttt{wangzengmao@whu.edu.cn} \\
  \And
  Di Wang \\
  Wuhan University \\
  \texttt{d$\_$wang@whu.edu.cn} \\
  \And
  Rong-Cheng Tu \\
  Nanyang Technological University \\
  \texttt{turongcheng@gmail.com} \\
  \And
  Bo Du \\
  Wuhan University \\
  \texttt{dubo@whu.edu.cn} \\
  \And
  Dacheng Tao \\
  Nanyang Technological University \\
  \texttt{dacheng.tao@ntu.edu.sg} \\
}

\begin{document}
\maketitle
\begin{abstract}
Reward models are critical for improving large language models (LLMs), particularly in reinforcement learning from human feedback (RLHF) or inference-time verification. Current reward modeling typically relies on scores of overall responses to learn the outcome rewards for the responses. However, since the response-level scores are coarse-grained supervision signals, the reward model struggles to identify the specific components within a response trajectory that truly correlate with the scores, leading to poor generalization on unseen responses.
In this paper, we propose to leverage generation probabilities to establish reward consistency between processes in the response trajectory, which allows the response-level supervisory signal to propagate across processes, thereby providing additional fine-grained signals for reward learning. Building on analysis under the Bayesian framework, we develop an intra-trajectory consistency regularization to enforce that adjacent processes with higher next-token generation probability maintain more consistent rewards. 
We apply the proposed regularization to the advanced outcome reward model, improving its performance on RewardBench. Besides, we show that the reward model trained with the proposed regularization induces better DPO-aligned policies and achieves better best-of-N (BON) inference-time verification results. Our code is provided in~\url{\detokenize{https://github.com/chaoyang101/ICRM}}.
\end{abstract}


\section{Introduction}

Reward models offer a quantitative measure of the quality of LLM responses based on human preferences or correctness, making them instrumental in improving LLM performance through RLHF~\citep{RLHF1,RLHF2,KTO,liu2025survey} or inference-time verification~\citep{GenRM,PAV}. In RLHF, reward models provide feedback signals that guide LLMs to generate desirable responses via reinforcement learning. In inference-time verification, they rank or filter outputs to ensure the selection of the most appropriate responses. Since these applications depend on reliable reward predictions for unseen responses, the generalization of the reward model is important~\citep{Generalization,GRM}.

To enhance the generalization of the reward model, extensive efforts have been made in the literature, including ensemble techniques~\citep{Ensemble,WARM}, data augmentation~\citep{data_aug,robustRM}, direct correction of bias caused by length~\citep{Length_control,ODIN}, and hidden-state regularization~\citep{GRM,contra}. Generally, these methods use holistic human evaluations of responses to learn the rewards of responses~\citep{BT-model,GRM,robustRM}. Despite their success, these models remain limited by coarse-grained supervision of the response-level scores, which hinders their ability to properly capture dependencies between responses and the processes. 
This may lead to overfitting to spurious features~\citep{GRM}, such as response length, instead of properly leveraging label-relevant components in the response trajectory, resulting in poor generalization to unseen responses. To identify content that influences the score of overall response, some approaches propose learning from process-level scores~\citep{prm,Math-Shepherd}. However, in many practical scenarios, obtaining such fine-grained annotations proves prohibitively expensive~\citep{Rest-mcts}.

To address these challenges, we propose establishing reward consistency between processes within the response trajectory, enabling response-level supervisory signals to propagate across processes and thereby enrich reward learning.  
Specifically, we utilize generation probabilities, which measure the likelihood of a generator producing subsequent sequences, to capture inter-process dependencies. Through Bayesian decomposition, we formalize the relationship between these generation probabilities and reward consistency: Rewards are more likely to become similar between processes in the same response when the generator assigns a higher probability to their sequential generation. 
Moreover, to prevent severe misjudgment of reward consistency caused by low generation probabilities, we focus on adjacent processes with minimal content variation. These process pairs often exhibit semantic continuity and, consequently, tend to have comparable rewards.

\begin{figure}[t]
\centering
\includegraphics[width=0.95\textwidth]{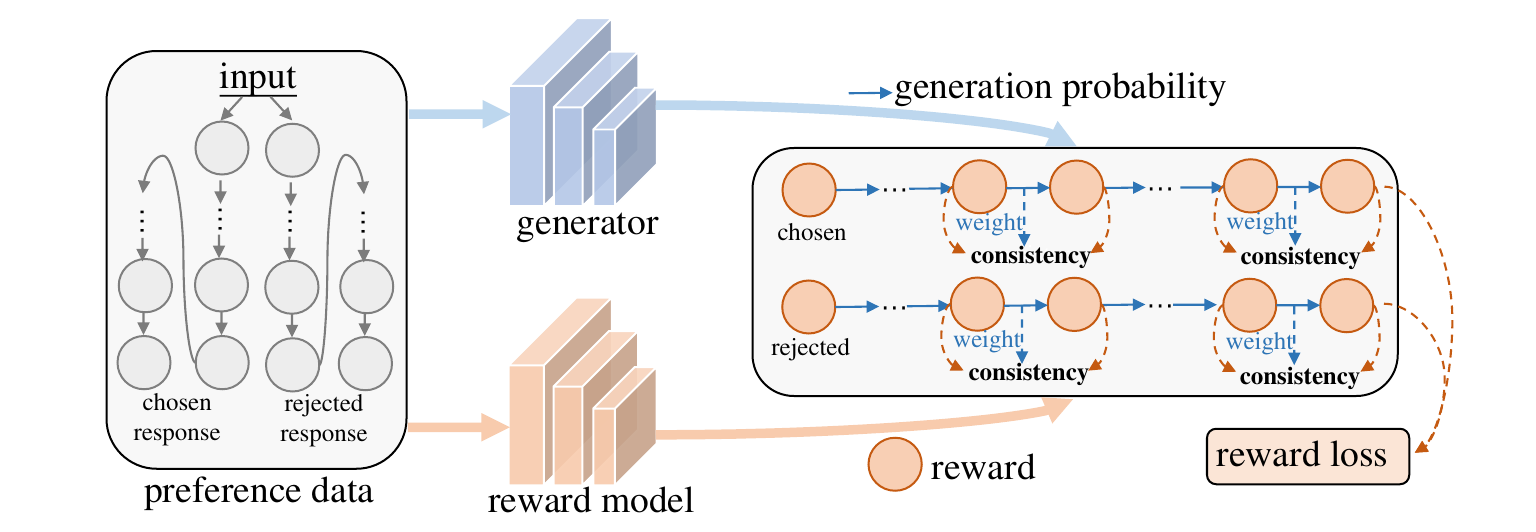}
\caption{Illustration of our proposed framework. Within this framework, while training the reward model to learn outcome rewards through standard reward loss, we introduce an additional intra-trajectory consistency regularization term. The regularization enforces stronger reward consistency between adjacent generation processes with higher next-token probabilities from the generator.}
\label{fig: motivation}
\end{figure}

To this end, we introduce intra-trajectory consistency regularization for reward modeling, termed ICRM. As shown in Figure~\ref{fig: motivation}, our framework consists of two components: a frozen generator that provides generation probabilities for regularization, and a reward model trained to predict outcome rewards. The reward model is regularized to produce more consistent rewards for adjacent processes with higher next-token generation probabilities. This design propagates response-level supervisory signals throughout the process trajectory, thereby improving generalization.

Finally, the main contributions can be summarized as follows:
\begin{itemize}
\item \textbf{Exploration.} We investigate the relationship between generation probabilities and reward consistency between processes through Bayesian decomposition.

\item \textbf{Method.} We propose a regularization method that enforces higher reward consistency between adjacent processes with higher next-token generation probabilities, thereby more effectively utilizing response-level supervisory signals for better generalization.

\item \textbf{Experiments.} We conduct extensive experiments to demonstrate that the proposed regularization improves the performance of the reward model in three evaluation tasks: standard reward modeling benchmarks, RLHF, and inference-time verification.
\end{itemize}

\section{Backgrounds}
\label{sec: problem definition}
Given an input prompt $x$, a standard language generator $\theta_g$, such as many current LLMs~\citep{gemma,llama3}, generates token sequences autoregressively, predicting each token $y_t$ conditioned on the preceding subsequence $y_{1:t-1}$ until reaching either a termination token or maximum length constraint. This process yields a complete output sequence $y = (y_1, \dots, y_n) = y_{1:n}$.
To enhance LLMs, many studies explore reinforcement learning training~\citep{rlhf3,GRPO} or employ inference-time verification~\citep{GenRM,PAV}. Both approaches require evaluating generated sequences, either through scoring or correctness assessment. This assessment is often referred to as the reward.

\textbf{Reward.} For an input $x$ with corresponding generated response sequence $y$ or a process $y_{1:k}$ consisting of the first $k$ tokens in $y$, reward functions can be categorized into two fundamental types: outcome reward $r(x,y)$ and process reward $r(x,y_{1:k})$. The outcome reward evaluates the complete response based on its final solution quality~\citep{process_benefit,GenRM}. In contrast, the process reward evaluates the scores of the intermediate processes within a response. Since it is not clear how to divide the processes of a common scenario, weesponse segment $y_{1:m}$ as a valid process. We should also note that in our definition, a full response can also be treated as a process. While this inclusive definition admits partial sentence fragments as independent processes, recent work in both RLHF~\citep{token-level_rlhf,token-level_rlhf2} and inference-time verification~\citep{GenARM} has demonstrated the empirical effectiveness of such fine-grained reward signals.

\textbf{Reward Modeling.} Many existing methods train reward models $\theta_r$ using overall response-level annotations. The current dominant approach for reward modeling is the Bradley-Terry model~\citep{BT-model}. For this model, the training dataset $D_{tr}$ whose unit is a triple $(x,y^w,y^l)$, where $x$ represents an input or a prompt, $y^w$ is a chosen response for $x$, and $y^l$ is a rejected response for $x$. To distinguish between the chosen response and the rejected response for a given input, we can optimize the Bradley-Terry reward model with the objective 
\begin{equation}
\begin{split}
& \mathcal{L}_{bt}= \mathbb{E}_{(x, y^w, y^l)\thicksim D_{tr}} \left[- \log \sigma (\theta_r(x, y^w)-\theta_r(x, y^l))\right],
\end{split}
\label{eq: bt_loss}
\end{equation}
where $\sigma$ is the sigmoid function. After optimization, the reward model can be used to provide outcome rewards for RLHF or inference-time verification.

\section{Method} 
\label{sec: method}
In this section, we introduce the intra-trajectory consistency regularization, which exploits the reward consistency relationship between processes in the response trajectory to propagate response-level supervisory signals. In what follows, we first discuss the connection between reward consistency between processes in a response trajectory and generation probability (cf. Section~\ref{sec: motivation}). Then, we introduce how this connection is implemented to regularize the reward model (cf. Section~\ref{sec: implementation}). Finally, we embed the proposed regularization term into the full training framework to learn more reliable outcome rewards for the full responses (cf. Section~\ref{sec: overall framework}).

\subsection{Establishment of Reward Consistency}
\label{sec: motivation}
Traditional reward modeling uses coarse response-level scores (e.g., pairwise preferences)~\citep{BT-model,GRM,robustRM}, making it difficult to assess fine-grained correctness~\citep{Fine_grained_human_feedback}. To introduce fine-grained signals, we propose establishing reward consistency relations between processes with the same response trajectory. This framework enables response-level supervisory signals to propagate throughout the trajectory, providing additional signals for reward learning. These derived signals help the model better capture contextual dependencies between processes. Besides, this is achieved without requiring additional manual annotation.

To establish the intra-trajectory consistency, we propose to leverage generation probabilities, the likelihood of a generator producing each subsequent sequence, to reflect the reward dependencies between processes. For example, research in the safety domain~\citep{safety} demonstrates that certain intermediate processes, such as phrases like "Sure, here is a detailed guide," often precede hazardous completions in response to harmful queries. Therefore, linking reward relationships between processes with their underlying generation probabilities is possible.

To achieve this objective, we formalize our approach by making key assumptions about the generation of the response. Specifically, we assume that responses are generated by a generator $\theta_g$ with an estimable conditional probability distribution, meaning each new token depends probabilistically on all previously generated tokens. This assumption aligns with the autoregressive nature of modern LLMs. Under these conditions, we can connect process $y_{1:m}$ and its subsequent process $y_{1:n}$  (where $m < n$)  for input $x$ with the generation probability through Bayesian decomposition:
\begin{equation}
P(e|x,y_{1:m}) = P(e|x,y_{1:n}) P(x, y_{1:n}|x,y_{1:m}) + \sum_{\bar{y}_{1:n} \in \bar{Y}_{1:n}} P(e|x,\bar{y}_{1:n}) P(x, \bar{y}_{1:n}|x,y_{1:m}),
\label{eq: decomposition}
\end{equation}
where $P(e|x, y)$ denotes the conditional probability of any event $e$ occurring given $(x,y_{1:m})$. $P(x, y_{1:n} \mid x, y_{1:m})$ represents the generation probability of sequence $y_{1:n}$ conditioned on $(x,y_{1:m})$, as computed by the generator. Since $y_{1:n}$ is the successor of $y_{1:m}$, $P(x, y_{1:n} \mid x, y_{1:m}) $ also equals to $ P(y_{m:n} \mid x, y_{1:m})$. $\bar{Y}_{1:n}$ denotes the set of all possible sequences with length $n$ excluding $y_{1:n}$. Following~\citep{PQM}, the reward $r(x,y)$ can be expressed as the probability that a response generated conditioned on $(x,y)$ achieves the maximum score. Then we can replace $P(e|x, y)$ with $r(x,y)$, so that the rewards are connected with the generation probability, represented as:
\begin{equation}
r(x,y_{1:m}) = r(x,y_{1:n}) P(x,y_{1:n}|x,y_{1:m}) + \sum_{\bar{y}_{1:n} \in \bar{Y}_{1:n}} r(x,\bar{y}_{1:n}) P(x,\bar{y}_{1:n}|x,y_{1:m}).
\label{eq: law_generation}
\end{equation}

From Eq.~\eqref{eq: law_generation}, since $\sum\limits_{\bar{y}_{1:n} \in \bar{Y}_{1:n}} P(\bar{y}_{1:n}|x,y_{1:m}) + P(y_{1:n}|x,y_{1:m}) = 1$, we observe that as the generation probability $P(x,y_{1:n}|x,y_{1:m})$ increases, the contribution of alternative completions $\bar{y}_{1:n}$ to the reward $r(x,y_{1:m})$ diminishes. Therefore, the reward $r(x,y_{1:m})$ becomes increasingly dominated by $r(x,y_{1:n})$, reducing the variance of the reward and leading to higher consistency between $r(x,y_{1:m})$ and $r(x,y_{1:n})$. Thus, generation probability and reward consistency can be linked.

\begin{remark}
It is worth noting that the Q-value~\citep{Math-Shepherd,PQM,PAV}, representing the expected benefit or probability of an action leading to a correct answer, is one of the interpretations of $r(x,y_{1:m})$ in Eq.~\eqref{eq: law_generation}. However, accurately estimating the Q-value of a process requires generating numerous complete responses and labeling them. Moreover, learning Q-values does not directly guarantee that the rewards between processes preserve the dependency in Eq.~\eqref{eq: law_generation}.
\end{remark}

The above analysis also implies an issue: When the generation probabilities between processes are low, reward similarity estimation may be unreliable. To address this, we incorporate reward consistency between semantically related processes. Inspired by text augmentation methods~\citep{word_drop2,word_dropout}, where partial masking preserves semantics, we assume that process semantics remain stable under limited suffix additions. This semantic invariance suggests correlated rewards between successive processes ($y_{1:m-1}$ and $y_{1:m}$). Consequently, we focus on generation probabilities and reward consistency for adjacent processes as a more tractable approach.

\subsection{Intra-Trajectory Consistency Regularization}
\label{sec: implementation}

Building on the above analysis, we propose \emph{intra-trajectory consistency regularization} to enforce more consistent rewards between adjacent processes with higher next-token generation probabilities. We subsequently present our reward formulation of processes under the Bradley-Terry framework, followed by the corresponding optimization objective.

For reward formulation, we note that standard reward outputs under the Bradley-Terry framework, $r(x,y_{1:m}) = \sigma(\theta_r(x,y_{1:m}))$, often saturate near the boundary values (0 or 1). To improve reward discriminability, we propose mean-centered calibration across preference triples $(x,y^w,y^l)$. 
For a process $y_{1:m}^w$ in the chosen response $y^w$, we define its calibrated reward as:
\begin{equation}
\hat{r}(x,y^w_{1:m}) = \sigma (\theta_r(x,y^w_{1:m}) - \frac{1}{|y^l|}\sum_{k=1}^{|y^l|}\theta_r(x,y^l_{1:k})),
\label{eq: chosen_reward}
\end{equation}
where $|y|$ denotes the length of sequence $y$. The mean value serves only as a calibration term and is excluded from gradient computation.
Analogously, for a process $y_{1:m}^l$ in the rejected response $y^l$, the calibrated reward is:
\begin{equation}
\hat{r}(x,y^l_{1:m}) = \sigma (\theta_r(x,y^l_{1:m}) - \frac{1}{|y^w|}\sum_{k=1}^{|y^w|}\theta_r(x,y^w_{1:k}) ).
\label{eq: rejected_reward}
\end{equation}

Building upon the calibrated rewards of processes, we can introduce a method to enforce reward consistency between adjacent processes. A direct method is to minimize calibrated reward distances (e.g., absolute differences) between adjacent processes. However, this method is ineffective under stochastic rewards (e.g., randomly initialized values), as forcing consistency between arbitrary rewards has limited meaning. Inspired by \citep{Fixmatch,Flexmatch}, we address this limitation through a mutually weighted binary cross-entropy loss that both learns semantically meaningful process rewards and promotes the reward consistency between adjacent processes.

Specifically, for a preference triple $(x, y^w, y^l)$, we assign process-level labels identical to the response preference label $s$, where $s=1$ indicates a chosen response $y^w$ and $s=0$ indicates a rejected one $y^l$. The weighting mechanism for a pair of adjacent processes $(y_{1:k-1}, y_{1:k})$ in a response combines two factors: (1) the next-token probability $P(x,y_{1:k}|x,y_{1:k-1}) = P(y_k|x,y_{1:k-1})= \theta_g(y_k \mid x, y_{1:k-1})$ from the generator $\theta_g$, and (2) the confidence of the calibrated reward of another paired process in predicting the preference label. Formally, for $y_{1:k-1}$ in the pair, its weight is computed as:
\begin{equation}
w(k \rightarrow k-1, s)=\theta_g(y_k|x,y_{1:k-1})\cdot (s\cdot\hat{r}(x,y_{1:k})+(1-s)\cdot (1-\hat{r}(x,y_{1:k}))).
\label{eq: weight}
\end{equation}
Similarly, for $y_{1:k}$ in the pair, its weight is formulated as: 
\begin{equation}
w(k-1 \rightarrow k, s)=\theta_g(y_k|x,y_{1:k-1})\cdot (s\cdot\hat{r}(x,y_{1:k-1})+(1-s)\cdot (1-\hat{r}(x,y_{1:k-1}))).
\label{eq: weight2}
\end{equation}
These weights are not used for gradient computation. When $s=1$, $y=y^w$. When $s=0$, $y=y^l$. Finally, let $\breve{r}(\cdot) = 1-\hat{r}(\cdot)$, the regularization loss for all training triples $(x, y^w, y^l)$ is:
\begin{equation}
\begin{split}
\mathcal{L}_{reg}= \mathbb{E}_{(x, y^w, y^l)\thicksim D_{tr}} &[-\sum_{k=2}^{|y^w|} w(k \rightarrow k-1, 1)\log \hat{r}(x,y^w_{1:k-1}) +  w(k-1 \rightarrow k, 1) \log \hat{r}(x,y^w_{1:k})  \\
& -\sum_{k=2}^{|y^l|} w(k \rightarrow k-1, 0) \log \breve{r}(x,y^l_{1:k-1}) + w(k-1 \rightarrow k, 0)  \log \breve{r}(x,y^l_{1:k})]. \\
\end{split}
\label{eq: reg_loss}
\end{equation}
In Eq.~\ref{eq: reg_loss}, the binary classification loss deviates the random process rewards to meaningful ones. Besides, since the losses of adjacent processes are mutually weighted by rewards from each other, their rewards can gradually become similar. The degree of this consistency constraint is implicitly governed by their next-token generation probabilities.
Consequently, Eq.~\ref{eq: reg_loss} prioritizes meaningful reward consistency for adjacent processes with higher next-token generation probabilities. We compare the L1 loss that minimizes absolute differences between process rewards in Appendix~\ref{sec: L1_evaluation}.

\subsection{Overall Training Framework}
\label{sec: overall framework}
In this section, we introduce how to incorporate intra-trajectory consistency regularization into the standard Bradley-Terry reward modeling to learn more reliable outcome rewards. 
Since existing reward modeling datasets are often aggregated from diverse sources, including some unavailable models, we begin by performing supervised fine-tuning (SFT) on a pre-trained language model to derive a generator $\theta_g$. This generator is optimized to align with the training dataset's generation probability distribution. Then this fine-tuned generator can provide next-token generation probability for computing $L_{reg}$. When the training data comes from a single known LLM, we can directly use that model as the generator. After acquiring the generator, we use two objectives to train the reward model: (1) a reward modeling loss $L_{bt}$ applied to the entire response, which facilitates learning the outcome rewards; and (2) a regularization $L_{reg}$ applied across processes. Finally, the overall loss to update the model is computed as:
\begin{equation}
\mathcal{L}_{toal}= (1-\alpha)\mathcal{L}_{bt} + \alpha \mathcal{L}_{reg} 
\label{eq: overall_loss}
\end{equation}
where $\alpha$ is a balance hyper-parameter. Compared to solely optimizing $\mathcal{L}_{bt}$, optimizing Eq.~\ref{eq: overall_loss} enables the model to focus on finer-grained signals. After training, the model can be used to provide outcome rewards. As an extension, we also explore an end-to-end variant where the reward model and generator share a backbone and are jointly optimized (see Appendix~\ref{sec: end-to-end}).

\section{Experiments}
\label{sec: experiment}

\subsection{Experimental Setup}
\label{sec: setup}
\textbf{Datasets.} (1) We train the reward models on the widely-used Unified-Feedback\footnote{https://huggingface.co/datasets/llm-blender/Unified-Feedback} and Skywork\footnote{https://huggingface.co/datasets/Skywork/Skywork-Reward-Preference-80K-v0.2}. The data of these two datasets are derived from multiple LLMs. Thus, to verify the proposed method in the situation where the generation distribution is known, we adopt Qwen2.5-7B-Instruct~\citep{qwen2} to generate 4 responses for each question in the training set of the prm-800k dataset~\citep{prm} and label these responses with gold answers. These generated responses constitute the Qwen-Generated dataset. (2) For experiments of RLHF, we sample about 500 prompts from the prompts of RewardBench as the test set and the rest as the training set.  (3) For experiments of inference-time verification, we evaluate the reward model on MATH-500~\citep{MAth500}, with BON datasets from~\citep{free_process}, containing processes generated by both Mistral-Instructor-v0.3~\citep{Mistral} and Llama-3-8B-Instruct~\citep{llama3}.

\textbf{Training Details.} (1) We train the proposed reward modeling method based on GRM~\citep{GRM}, which achieves competitive results in the RewardBench benchmark. The resulting reward model is termed \textbf{ICRM}. We validate the proposed method on Gemma-2B-it~\citep{gemma}, Llama3-8B-instruct~\citep{llama3}, and Qwen-1.5B-Instruct. The hyperparameter $\alpha$ is set to 0.1 in all of our experiments. Additional training configurations, including the procedure for training Qwen-1.5B-Instruct with unpaired data, are detailed in Appendix~\ref{sec: implementation details}. (2) For RLHF, we adopt DPO strategy~\citep{DPO} to train the policy model with Gemma-2B-it as the initial model. Following~\citep{Rlhf_workflow,robustRM}, we generate 8 responses for each prompt in the training set of RLHF. Then these responses are scored by the 2B reward model trained with Unified-Feedback 400K. The best-worst response pairs for each prompt are used to train the DPO policy. All experiments are implemented on at most two A800 GPUs, each with 80GB of memory.

\begin{table}[t]
\caption{Accuracy results on RewardBench with different sizes of training samples from Unified-Feedback. The base model is Gemma-2B-it. The best results in a column of a series are highlighted in bold. * indicates that the result is copied from~\citep{GRM}. }
\centering
\resizebox{0.9\linewidth}{!}{
\begin{tabular}{c|c c c c c}
\toprule[1.5pt]
 Reward Model  &  Chat &  Chat-Hard & Safety & Reasoning & Average \\
\midrule[1pt]
\multicolumn{6}{c}{40K training samples} \\
\midrule[1pt]
Classifier*  & 95.8 & 37.3 & 59.9 & 64.8 & 64.5 \\
Classifier + margin* & \textbf{97.2} & 37.5 & 56.8 & 72.7  & 66.1 \\
Classifier + label smooth*  & 91.6 & 39.0 & 53.8 & 60.2 & 61.1\\
Classifier + Ensemble*  & 96.1 & 38.2 & 58.8 & 67.6 & 65.2\\
GRM*  & 94.7 & 40.8 & 65.4 & \textbf{77.0} & 69.5  \\
GRM (reproduced)   & 96.6 & 41.0  &  80.8 & 73.8 & 73.0  \\
ICRM (Ours)  &  95.0 & \textbf{48.2} & \textbf{84.2} &  75.8 & \textbf{75.8} \\
\midrule[1pt]
\multicolumn{6}{c}{400K training samples} \\
\midrule[1pt]
Classifier*  & 89.7 & 50.7 & 74.7 & 57.9 & 68.2 \\
Classifier + margin*  & 89.7 & 47.1 & 70.7 & 43.6 & 62.8 \\
Classifier + label smooth*  & 94.1 & 47.1 & 67.5 & 79.7 & 72.1 \\
Classifier + Ensemble*  & 89.6 & \textbf{50.2} & 72.7 & 59.0 & 69.3 \\
GRM*  & \textbf{96.1} & 40.1 & 80.3 & 69.3  & 71.5 \\
GRM (reproduced)  & 95.3 & 43.2  &  78.9 &  75.2 & 73.2   \\
ICRM (Ours)  & 95.5 & 44.5 & \textbf{84.5} & \textbf{78.2} & \textbf{75.7}  \\
\bottomrule[1.5pt]   
\end{tabular}}
\label{table: Unified-Feedback}
\end{table}

\begin{table}[t]
\caption{Accuracy results on RewardBench with training data from Skywork+Unified-Feedback 40K and  Llama3-8B-instruct. Best results is highlighted in bold. ``avg'' refers to the use of an exponential moving average (EMA) of the rewards from the trailing tokens during inference, with the smoothing applied backward from the last token and a decay factor of 0.5.}
\centering
\resizebox{0.8\linewidth}{!}{
\begin{tabular}{c|c c c c c}
\toprule[1.5pt]
Reward Model &  Chat &  Chat-Hard & Safety & Reasoning & Average \\
\midrule[1pt]
GRM & 95.5 & 74.1 & 86.6 & 89.0 & 86.3 \\
GRM-avg & \textbf{96.9} & 74.1 & 85.0 & 91.2 & 86.8 \\
ICRM (Ours) & 95.2 & 75.9 & 86.2 & 89.7 & 86.8 \\
ICRM-avg & 96.1 & \textbf{78.1} & \textbf{87.3} & \textbf{95.0} & \textbf{89.1} \\
\bottomrule[1.5pt]   
\end{tabular}}
\label{table: skywork}
\end{table}

\textbf{Baselines and Evaluation Details.} In this paper, we evaluate the proposed method in three tasks: the standard RewardBench benchmark~\citep{Rewardbench}, RLHF, and inference-time verification. (1) For RewardBench benchmark, we consider Classifier trained with Eq.~\ref{eq: bt_loss}, Classifier+Margin~\citep{LLAMA2}, Classifier+Label Smooth~\citep{label_smooth}, Classifier+Ensemble~\citep{reward_ensembles}, and GRM~\citep{GRM} as baselines. (2) For RLHF, inspired by~\citep{alpha,GRM}, to avoid the high costs of human evaluation, we employ the QRM-Llama3.1-8B model~\footnote{https://huggingface.co/nicolinho/QRM-Llama3.1-8B-v2} as a gold scoring model according to its strong performance on the RewardBench benchmark. The trained DPO model generates responses for the prompt in the test set by greedy sampling. (3) When evaluating inference-time verification, we adopt the best-of-N (BON) metric, which measures the probability that the response selected by the reward model from N alternative responses is the correct answer or the best solution.

\subsection{Evaluation on Reward Modeling Benchmark}

\textbf{Results on ReardBench Benchmark.} In Table~\ref{table: Unified-Feedback} and Table~\ref{table: skywork}, we evaluate different reward models on the RewardBench benchmark across four categories: Chat, Chat-Hard, Safety, and Reasoning. The results demonstrate that our proposed method achieves higher average scores than all baselines trained on the same dataset and model architecture. Furthermore, the proposed method achieves better results than the baseline in most categories. These results demonstrate that the proposed regularization enhances the generalization capability of the reward model.

\textbf{Comparison of Different Sizes of Training Samples.} Following~\citep{GRM}, we also compare the performance of the proposed method under different sizes of training samples. In Table~\ref{table: Unified-Feedback}, we present results for both 40K and 400K training samples. Additional results for other training sizes are provided in Appendix~\ref{sec: datasize}. These results demonstrate that the proposed method consistently outperforms the baseline GRM in terms of average score across varying amounts of training data. This highlights the robustness of the improvements of the proposed method under different data sizes.

\textbf{Utilization of the Process Rewards.} The practice of evaluating response correctness through process rewards is well-established in mathematical reasoning~\citep{prm,PQM}. Since our method also utilizes process rewards, we examine their impact. To integrate process rewards, we compute an exponentially weighted moving average of rewards, starting from the final token of each response and proceeding backward through the sequence. Given a response with length $m$ and the average decay $d$, the average reward $r=\sum_{i=1}^m d^{m-i} r(x,y_{1:i})$. The results are shown in Table~\ref{table: skywork}. Our findings reveal that incorporating this strategy leads to improvements for both GRM and our method in the reasoning group, with occasional gains in other groups. Crucially, our method consistently surpasses GRM when using this approach. These results suggest that our approach learns reliable process rewards to some extent.

\begin{table}[t]
\caption{Results of DPO policy with guidance from different reward models. ``Gold score'' is the normalized average score acquired from the gold scoring model. ``Win ratio'', ``Tie ratio'' and ``Lose ratio'' are obtained by taking each other as comparison model. The ``Win ratio'', ``Tie ratio'' and ``Lose ratio'' represent the proportions of comparisons in which a model's outputs are preferred (win), deemed equivalent (tie), or dispreferred (lose) relative to another model's outputs.}
\centering
\resizebox{0.8\linewidth}{!}{
\begin{tabular}{c|c c c c}
\toprule[1.5pt]
Reward Model &  Gold score$\uparrow$ & Win ratio$\uparrow$ & Tie ratio & Lose ratio$\downarrow$ \\
\midrule[1pt]
GRM & 0.676 & 47.3 & 2.1 & 50.6 \\
ICRM (Ours) & 0.678 & 50.6 & 2.1 & 47.3 \\
\bottomrule[1.5pt]   
\end{tabular}}
\label{table: dpo_aligned}
\end{table}

\subsection{Evaluation on RLHF}
We evaluate reward model quality by analyzing the performance of policies optimized under different reward functions, with results presented in Table~\ref{table: dpo_aligned}. Our analysis reveals two findings: (1) the policy induced by the proposed reward model achieves higher gold scores compared to the baseline GRM, and (2) demonstrates superior prompt-conditional response quality, generating higher-scoring outputs for identical prompts. These results collectively validate the enhanced capability of our reward model in RLHF pipelines, particularly in its ability to guide policy to generate more desirable responses.

\begin{table}[t]
\caption{Best-of-N (BON) inference-time verification results of responses from different polices and pass@N. All reward models are implemented using Qwen-1.5B-Instruct as the base model and trained on the Qwen-Generated dataset.}
\centering
\resizebox{1.0\linewidth}{!}{
\begin{tabular}{c c | c c c c c}
\toprule[1.5pt]
Policy & Reward Model &  Pass@2 &  Pass@4 &  Pass@8 & Pass@16 & Average \\
\midrule[1pt]
\multirow{2}{*}{Mistral-Instructor-v0.3} & GRM & 11.8 & 11.8 & 12.6 & 14.2 & 12.6 \\
& ICRM (Ours) & 11.8 & 12.8 & 14.6 & 14.0 & 13.3 \\
\midrule[1pt]
\multirow{2}{*}{Llama-3-8B-Instruct} & GRM & 45.6 & 46.8 & 49.2 & 45.0 & 46.7 \\
& ICRM (Ours) & 45.8 & 47.2 & 51.2 & 50.0 &  48.6 \\
\bottomrule[1.5pt]   
\end{tabular}}
\label{table: BON_math}
\end{table}

\begin{table}[t]
\caption{Ablation study for the proposed regularization. ``w/o adjacent reg'' means that the reward of another process is not used for weighting in the proposed regularization. ``w/o generation reg'' means that generation probability is not used in the proposed regularization. Training dataset is 40K samples from Unified-Feedback, and model is Gemma-2B-it.}
\centering
\resizebox{0.85\linewidth}{!}{
\begin{tabular}{c|c c c c c}
\toprule[1.5pt]
Method  &  Chat &  Chat-Hard & Safety & Reasoning & Average \\
\midrule[1pt]
w/o adjacent reg & 96.1 & 43.2 & 80.8 & 74.0 & 73.5 \\
w/o generation reg & 95.2 & 46.9 & 83.5 & 75.2 & 75.2 \\
\midrule[1pt]
Overall & 95.0 & 48.2 & 84.2 &  75.8 & 75.8 \\
\bottomrule[1.5pt]   
\end{tabular}}
\label{table: ablation study}
\end{table}

\begin{figure}[t]
\centering
\subfigure[Best of 4 responses]{
    \includegraphics[width=0.33\linewidth]{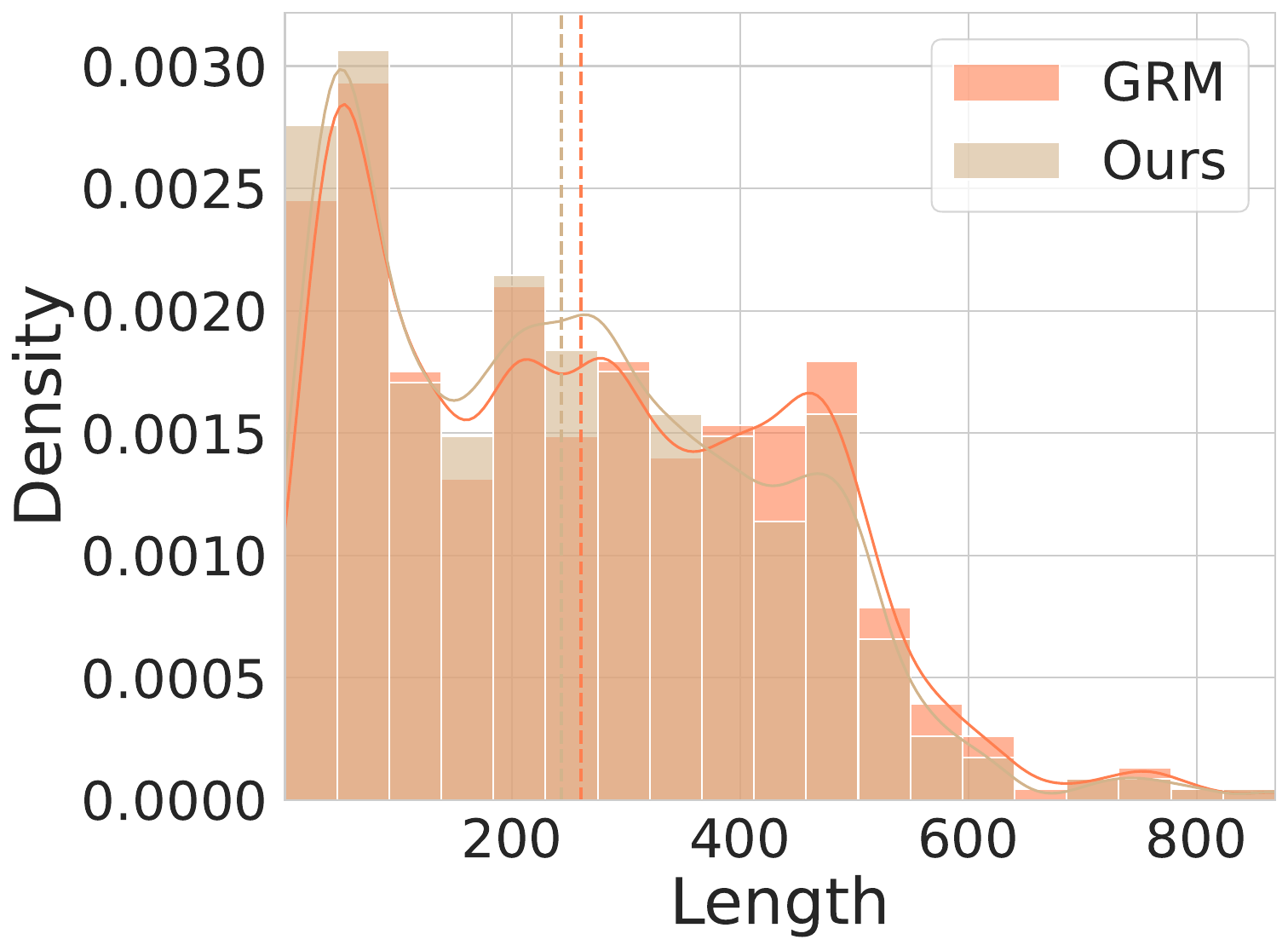} 
  }
\hspace{-0.03\linewidth}
\subfigure[Best of 16 responses]{
    \includegraphics[width=0.33\linewidth]{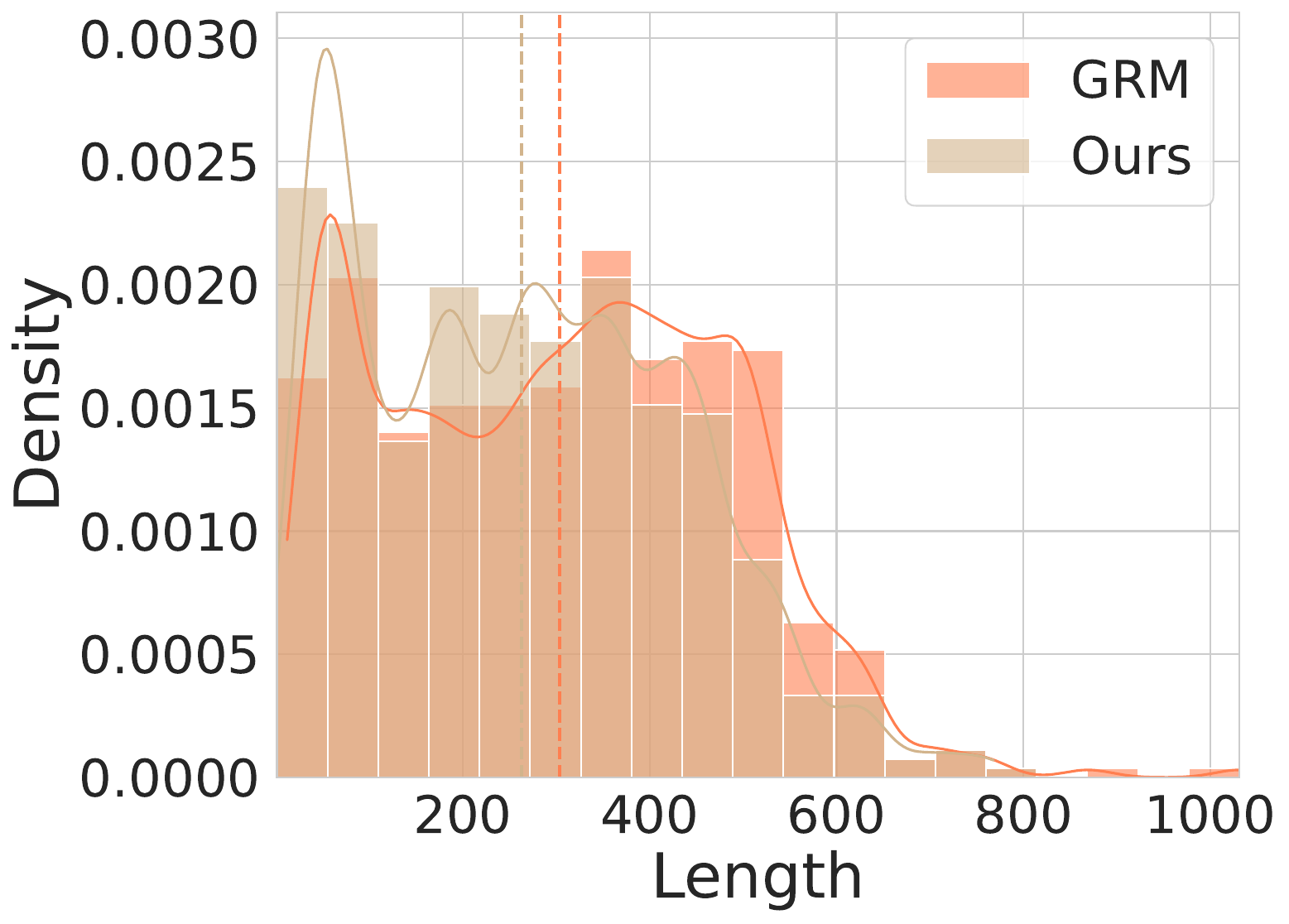} 
  }
\hspace{-0.03\linewidth}
\subfigure[DPO policy]{
    \includegraphics[width=0.33\linewidth]{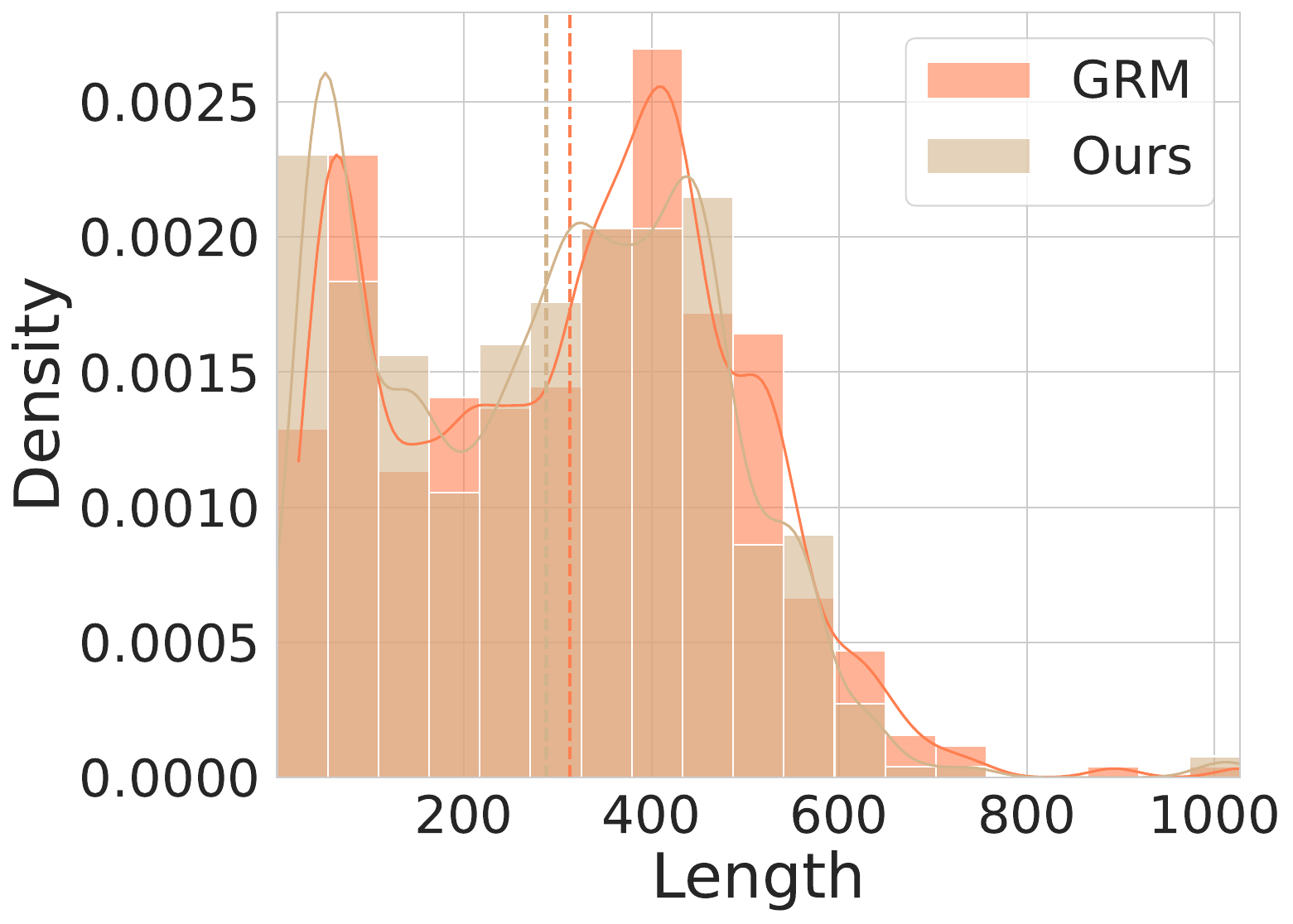} 
  }
\caption{Length distribution of response of various policies induced by GRM and ICRM. Average length is marked by the dashed line. Prompts for generation are derived from the test set in the RLHF experiments. Gnenration model is Gemma-2b-it.}
\label{fig: length}
\end{figure}

\subsection{Evaluation on Inference-Time Verification}
To demonstrate inference-time verification capability of the reward models, we evaluate their performance in Best-of-N (BON) sampling scenarios. As shown in Table~\ref{table: BON_math}, our method achieves superior mean accuracy compared to the baseline GRM when applied to both Mistral-Instructor-v0.3 and Llama-3-8B-Instruct, with particularly notable gains on Llama-3-8B-Instruct. The proposed approach outperforms the baseline in most cases, validating its effectiveness for inference-time verification. Importantly, when training our reward model used in this evaluation, the LLM used to generate the training data is directly used as the generator. Thus, the results also confirm our method's ability to leverage accessible training data generators for performance improvement.

\subsection{Empirical Analysis}
\textbf{Ablation Study.} The proposed intra-trajectory consistency regularization loss incorporates two key weighting components in Eq.~\eqref{eq: weight} and Eq.~\eqref{eq: weight2} : (1) weights derived from rewards of adjacent processes, and (2) weights based on generation probabilities. To evaluate their respective contributions, we conduct ablation studies in Table~\ref{table: ablation study}, where each component is removed individually.
The results demonstrate that removing either weighting component typically degrades performance, with the weights from rewards of adjacent processes showing a more substantial impact. These findings underscore the importance of both weighting mechanisms in the proposed intra-trajectory reward consistency regularization term.

\textbf{Length Analysis.} Length represents a common superficial feature that reward models may exploit, often manifesting as a preference for longer responses~\citep{length}. To investigate this length bias, we analyze response length distributions across different policy outputs (Figure~\ref{fig: length}). Our results demonstrate that policies optimized under our reward model consistently produce shorter responses than the baseline. This finding indicates that the proposed method has the potential to reduce the model's reliance on length as a proxy for response quality.

\begin{figure}[t]
\centering
\subfigure[A chosen response]{
    \includegraphics[width=0.47\linewidth]{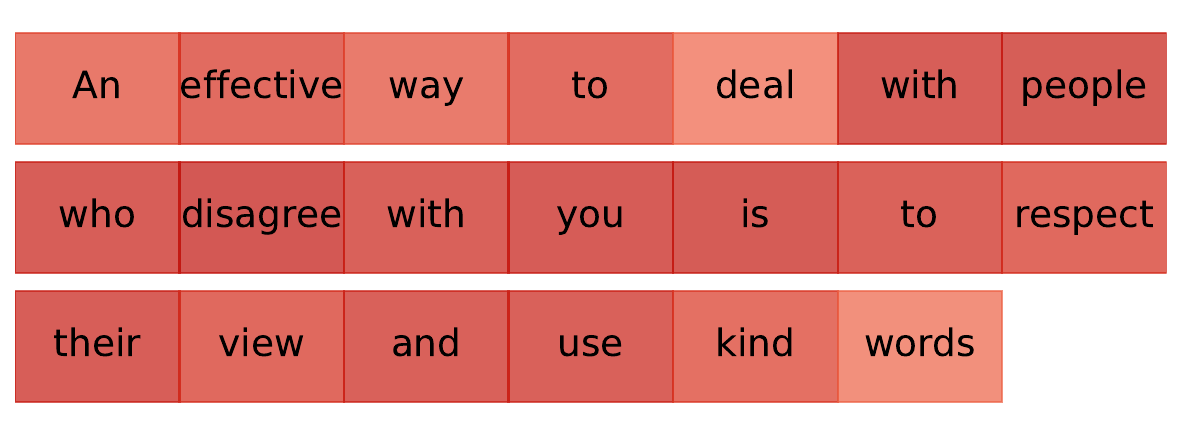} 
  }
\hspace{-0.03\linewidth}
\subfigure[A rejected response]{
    \includegraphics[width=0.47\linewidth]{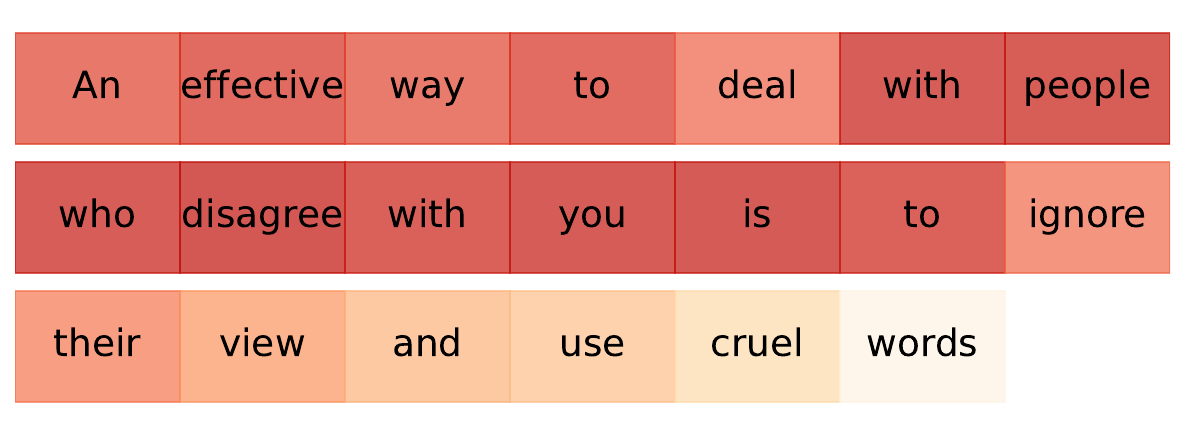} 
  }
\caption{Heatmap of the rewards acquired by ICRM for different processes, in which the reward of a process is shown in the last token of the process, and darker colors indicate higher rewards. The prompt of the two responses is ``What is an effective way to deal with people who disagree with me?''. The left response (with ``respect'' and ``kind'') is preferred as the chosen response over the right (with ``ignore'' and ``cruel'') due to its more positive wording concerning the context.}
\label{fig: process reward}
\end{figure}

\textbf{Visualization of Process Rewards.} To elucidate the relationship between process-level rewards and response labeling learned by the proposed method, we visualize reward trajectories for two minimally contrasting responses that differ only in a few critical words (Figure~\ref{fig: process reward}). Our analysis reveals that: (1) the model systematically assigns higher rewards to processes containing semantically favorable words in context, and (2) for rejected responses, rewards exhibit gradual degradation rather than immediate drops, for instance, the token "ignore" triggers only a mild penalty, while subsequent unfriendly terms like "cruel" induce sharper declines. This demonstrates our method's capacity to develop nuanced and context-sensitive reward signals at the process level.

\section{Conclusion and Discussion of Limitations}
\label{sec: conclusion}

\textbf{Conclusion.} In this paper, we aim to mitigate the limitation in conventional reward modeling: the inherent coarseness of response-level supervision, which fails to capture fine-grained, reward-determining components within the generated response trajectory. To mitigate this issue, we propose a novel intra-trajectory consistency regularization to provide additional fine-grained signals for reward learning based on response-level supervision. Grounded in a Bayesian framework, the proposed regularization enforces that adjacent processes with higher next-token generation probabilities maintain more consistent rewards. Experimental results on the RewardBench benchmark, RLHF, and inference-time verification validate the effectiveness of the proposed regularization in improving the advanced reward model.

\textbf{Limitations}. The proposed method requires a generator to provide generation probabilities, which introduces two computational overheads: (1) forward passes through the generator during training of the reward model, and (2) potential generator fine-tuning.
For the first overhead, the measured training speeds in Appendix~\ref{sec: implementation details} show that the additional computational cost is not significant. Besides, the time cost can be further reduced through pre-processing with a larger batch size.
The second overhead can be avoided when all training data comes from a single white-box generator, a not uncommon scenario in RLHF. Due to resource constraints, we limit our validation to smaller models and employ model-based evaluation to assess the response quality of different policies. Scaling to larger models and developing more accurate validation methods remain important future directions.

\bibliographystyle{unsrtnat}
\bibliography{references} 

\newpage

\appendix

\section{Related Works}
\label{sec: related work}

To begin with, we discuss the related works to this study.

\textbf{Generalization of Reward Models.} The generalization of reward models to unseen responses is essential for improving their robustness in RLHF and inference-time verification~\citep{Generalization,GRM}. To improve it, multiple approaches have been developed, such as ensemble techniques~\citep{esm3,WARM} , data augmentation~\citep{data_aug,robustRM}, direct correction of measurable bias~\citep{Length_control,ODIN}, and hidden-state regularization~\citep{GRM}. For example,~\citep{Ensemble} proposes to learn multiple estimators and combine them to improve the robustness of the rewards.~\citep{robustRM} introduces a data augmentation approach derived from the causal framework to differentiate between contextual signals and context-free artifacts.~\citep {ODIN} proposes a framework trained to predict both rewards and lengths so that it can disentangle the representation of the content quality from the lengths of responses.~\citep{GRM} introduces SFT and DPO losses to regularize the hidden states of reward models.
While these methods strengthen RMs for AI alignment, their effectiveness remains constrained by sparse response-level supervision, limiting further generalization.

\textbf{Outcome Rewards and Process Rewards.} 
Outcome rewards and process rewards represent two fundamental reward paradigms in reinforcement learning and AI alignment~\citep{PPO,orm2,process_benefit}. Outcome rewards, which evaluate final results (e.g., task completion), are simple to define and require minimal annotation effort. However, they often fail to guide desirable behaviors in complex, long-horizon tasks~\citep{prm}. In contrast, process rewards, particularly critical in autoregressive generation, capture causal dependencies between tokens by evaluating intermediate steps (e.g., reasoning coherence), as demonstrated by~\citep{PQM}. While effective, process reward learning typically demands costly step-level annotations~\citep{Math-Shepherd,process_label2}. Recent work mitigates this through automated methods, such as~\citep{Math-Shepherd}'s stepwise scoring based on downstream answer correctness. Yet, such approaches remain suboptimal without a free annotator, leaving response-level annotation the prevailing standard for reward modeling. Therefore, it is meaningful to explore methods for incorporating more process-related signals based solely on response-level labeling.

\section{Implementation Details}
\label{sec: implementation details}
In this section, we provide more implementation details.

\textbf{Overall Information.} Our code is based on LlamaFactory~\citep{LlamaFactory}, a powerful code framework that supports multiple LLMs training strategies. The version number we adopted is 0.9.1.dev0. The LLMs involved in the experiments are all from HuggingFace. In the experiments, the models are trained based on LoRA~\citep{LoRA}. We employ the DeepSpeed framework\footnote{https://github.com/deepspeedai/DeepSpeed?tab=readme-ov-file} to enhance training efficiency and reduce GPU memory requirements. For models less than 2B parameters, we utilize the ZeRO-2 offload configuration. When training includes 8B models, we adopt the ZeRO-3 offload setting.

\textbf{More Details for Training of Reward Models.} Our reward model training procedure primarily follows GRM~\citep{GRM}, specifically adopting their GRM-sft configuration. GRM-sft incorporates an additional SFT loss alongside the standard reward loss to regularize hidden-state representations, with a default weighting factor of 0.001. The reward model is initialized using the backbone of a pre-trained LLM with an additional randomly initialized linear logistic regression layer (dropout=0.1). Common hyper-parameters are detailed in Table~\ref{table: reward model hyper-paramaters}.
Key variations include batch size and training epochs: for datasets less than 40K, we use batch size 12 with 2 epochs; other experiments employ batch size 24 with 1 epoch. Notably, our proposed method occasionally requires a generator during reward model training. The generator for the 2B model is trained with 40K samples from the Unified-Feedback dataset. The generator for the 8B model is trained with 40K samples from the Unified-Feedback dataset and the Skywork dataset. When training, it shares the same configuration as the reward model (Table~\ref{table: reward model hyper-paramaters}), except for a reduced learning rate (1e-7) and a single epoch to prevent overfitting.
For the reward model that is only trained with the Unified-Feedback dataset, we set the sequence cutoff length to 1,024 tokens. All other reward models and generators use a 2,048-token cutoff to better handle long-form data.

\begin{table}[t]
\caption{Common hyper-parameters in the experiments.}
\centering
\resizebox{0.6\linewidth}{!}{
\begin{tabular}{c @{\hspace{1.5cm}} c}
\toprule[1.5pt]
Quantization for training & bf16 \\
LoRA r & 32 \\
LoRA alpha & 64 \\
Optimizer & Adamw \\
Learning rate & 1e-5 \\
Learning rate scheduler & cosine \\
Warmup ratio & 0.03 \\
\bottomrule[1.5pt]   
\end{tabular}}
\label{table: reward model hyper-paramaters}
\end{table}

\textbf{Hyper-Parameters for RLHF.} In RLHF experiments, the process consists of two key components: response generation and subsequent training using these generated responses. For data generation, we employ a temperature of 1.0 and top-p sampling (p=1.0), producing 8 responses per prompt with a maximum length of 1,536 tokens. These responses are then evaluated by our 2B reward model trained on the Unified-Feedback 400K dataset.
We select the best-worst response pairs from each prompt for DPO policy training. The DPO configuration largely follows the settings in Table~\ref{table: reward model hyper-paramaters}, with two modifications: (1) a learning rate of 5e-6, and (2) the addition of a 0.1 scaling factor for comparing current and reference model probabilities. All DPO experiments run for 2 epochs.

\textbf{Computational Resources.} All experiments are implemented in at most two NVIDIA RTX A800 GPUs, which have about 160G memory. During training of the 2B GRM reward model, the baseline in this paper, with a batch size of 12, we observe a training speed of 4.6 seconds per iteration. Under identical batch size conditions, the 2B reward model trained with the proposed method achieves a comparable training speed of approximately 5.4 seconds per iteration.

\textbf{Implementation for Reward Modeling with Unpaired Data.} When paired data are not available, discriminative reward modeling is usually used to learn the reward. The training dataset $D_{tr}$ consists of triples $\{x^i,y^i,c^i\}_{i=1}^N$, where $x^i$ represents input for $i^{th}$ example, $y^i$ is a response for $x^i$, and $c^i$ is a gold binary classification label for $x^i$. To estimate the quality of a response for an input, we can train a discriminative reward model $\theta_r$ with the binary cross-entropy loss, namely, 
\begin{equation}
\mathcal{L}_{d}= \mathbb{E}_{(x, y, c) \thicksim D_{tr}} [- c \cdot \log(\sigma(\theta_r(x,y))) - (1-c)\cdot \log(1 - \sigma(\theta_r(x,y)))].
\label{eq: dis_loss}
\end{equation}
In this case, when computing the proposed regularization term, no calibration is required for the reward, i.e, $\hat{r}(x,y)=\sigma(\theta_r(x,y))$ in Eq.~\eqref{eq: reg_loss}. Finally, based on discriminative reward modeling, we only need to add $\mathcal{L}_{d}$ and the proposed regularization term, similar to Eq.~\eqref{eq: overall_loss}.

\section{Additional Experimental Results}
\label{sec: additional experimental results}

To further support our methods, we present additional experimental results that extend beyond those reported in the main paper. These include analyses of model performance across varying training sample sizes, an expanded evaluation of process reward utilization, and results obtained under an end-to-end training framework. We also provide BON evaluations on non-mathematical tasks, along with an assessment of reward alignment using L1 distance.

\subsection{Results on Different Sizes of Training Samples}
\label{sec: datasize}
\begin{table}[t]
\caption{Accuracy results on RewardBench with different sizes of training samples from the Unified-Feedback dataset. The base model is Gemma-2b-it.}
\centering
\resizebox{0.7\linewidth}{!}{
\begin{tabular}{c|c c c c c}
\toprule[1.5pt]
Reward Model &  4K &  10K & 40K & 400K & Average \\
\midrule[1pt]
GRM & 59.5 & 64.1 & 73.0 & 73.2 & 67.5 \\
ICRM (Ours) & 61.3 & 64.3 & 75.8 & 75.7 & 69.3 \\
\bottomrule[1.5pt]   
\end{tabular}}
\label{table: data_size}
\end{table}

We further investigate the performance of the proposed method on different training data scales.  We show the accuracy results on RewardBench with different sizes of training samples from the Unified-Feedback dataset in Table~\ref{table: data_size}. The experimental results demonstrate that the reward model trained with the proposed method consistently outperforms the baseline GRM method, validating its stability across varying amounts of training data.

\subsection{More Results on Utilization of Process Rewards}
\label{sec: tail average}
\begin{table}[t]
\caption{Accuracy results on RewardBench with training data from Skywork+Unified-Feedback 40K and  Llama3-8B-instruct. ``avg-val'' refers to the use of an exponential moving average (EMA) of the rewards from the trailing tokens during inference, with the smoothing applied backward from the last token and a decay factor of ``val''. Best results are highlighted in bold.}
\centering
\resizebox{0.8\linewidth}{!}{
\begin{tabular}{c|c c c c c}
\toprule[1.5pt]
Reward Model &  Chat &  Chat-Hard & Safety & Reasoning & Average \\
\midrule[1pt]
GRM & 95.5 & 74.1 & 86.6 & 89.0 & 86.3 \\
GRM-avg-0.5 & 96.9 & 74.1 & 85.0 & 91.2 & 86.8 \\
GRM-avg-0.7 & \textbf{97.2} & 70.6 & 81.9 & 90.1
 & 84.5 \\
GRM-avg-0.9 & 84.9 & 65.8 & 70.8 & 82.6 & 76.0
 \\
ICRM (Ours) & 95.2 & 75.9 & 86.2 & 89.7 & 86.8 \\
ICRM-avg-0.5 & 96.1 & \textbf{78.1} & 87.3 & 95.0 & \textbf{89.1} \\
ICRM-avg-0.7 & 93.3 & 76.7 & \textbf{88.4} & \textbf{96.0}
 & 88.6 \\
ICRM-avg-0.9 & 90.8 & 73.2 & 88.0 & 95.7
& 86.9 \\
\bottomrule[1.5pt]   
\end{tabular}}
\label{table: average_more}
\end{table}

We further investigate the effectiveness of process rewards by analyzing performance under an exponential moving average (EMA) of process rewards from trailing tokens. Table~\ref{table: average_more} compares different decay rates when averaging rewards, focusing on their impact relative to the final token's reward.
While over-reliance on process rewards can degrade performance, our method consistently outperforms GRM under this approach and exhibits stronger robustness against performance drops. This suggests that the learned process rewards have the potential to capture aspects of the overall response quality.

\subsection{Results under End-to-End Training Framework}
\label{sec: end-to-end}

\begin{table}[t]
\caption{Average accuracy results on RewardBench with different sizes of training samples from the Unified-Feedback dataset under different training settings. ``Training-time generator'' represents an end-to-end variant where the reward model and generator share a backbone and are jointly optimized. ``Pre-learned generator'' represents a two-stage variant where the generator is learned before the training of the reward model. }
\centering
\resizebox{0.7\linewidth}{!}{
\begin{tabular}{c|c c c c}
\toprule[1.5pt]
Standard &  4K &  10K & 40K & Average \\
\midrule[1pt]
Training-time generator & 61.3 & 64.4 &  74.7 & 66.8 \\
Pre-learned generator & 61.3 & 64.3 & 75.8 & 67.1 \\
\bottomrule[1.5pt]   
\end{tabular}}
\label{table: end_to_end}
\end{table}

To mitigate the inevitable computational overhead introduced by using a separate generator, we investigate a more efficient end-to-end training approach that jointly trains both the reward model and generator on a shared backbone network. Specifically, our architecture features: (1) a linear reward head forming the reward model $\theta_r$, and (2) a parallel linear generation head forming the generator $\theta_g$, both attached to the same backbone. The total training loss combines the reward optimization loss (Eq.~\eqref{eq: overall_loss}) with the generator's SFT loss, where we prevent model perturbation by zeroing out backpropagated gradients to the backbone model from the SFT loss. As shown in Table~\ref{table: end_to_end}, this end-to-end approach maintains reasonable accuracy with limited training data but exhibits degraded performance at larger scales. These results suggest the potential of the end-to-end approach to maintain training efficiency without significant performance loss on a low data scale.

\subsection{BON Results Beyond Math Tasks}
\label{sec: BON_general}
\begin{table}[t]
\caption{Best-of-8 results of different policy induced different reward models. Prompts are acquired from the test data in the RLHF experiments. The reward models are trained from 400K samples from Unified-Feedback with Gemma-2b-it as the base model. ``Win ratio'', ``Tie ratio'', and ``Lose ratio'' are obtained by taking the methods of comparison to each other as the baseline. The ``Win ratio'', ``Tie ratio'' and ``Lose ratio'' represent the proportions of comparisons in which a model's outputs are preferred (win), deemed equivalent (tie), or dispreferred (lose) relative to another model's outputs.}
\centering
\resizebox{0.8\linewidth}{!}{
\begin{tabular}{c c| c c c}
\toprule[1.5pt]
Policy & Reward Model & Win ratio$\uparrow$ & Tie ratio & Lose ratio$\downarrow$ \\
\midrule[1pt]
\multirow{2}{*}{Gemma-2b-it} & GRM  & 18.6  & 62.0 & 19.4\\
& ICRM (Ours)  & 19.4 & 62.0 & 18.6 \\
\midrule[1pt]
\multirow{2}{*}{Llama3-8B-instruct} & GRM  & 15.0 & 65.2 & 19.8 \\
& ICRM (Ours)  & 19.8 & 65.2 &  15.0 \\
\bottomrule[1.5pt]   
\end{tabular}}
\label{table: BON_general}
\end{table}

We conduct a systematic evaluation to assess the efficacy of the proposed method in enhancing inference-time verification capabilities across general scenarios. As detailed in Table~\ref{table: BON_general}, we present comparative Best-of-8 results for policies derived from distinct reward models. The experimental results reveal two findings: (1) our method consistently outperforms baseline approaches on both the 2B and 8B policy scales, and (2) the performance advantage becomes more pronounced with the 8B policy. These empirical results further validate that the proposed method effectively improves inference-time verification performance in practical applications.

\subsection{Evaluation with L1 Reward Alignment}
\label{sec: L1_evaluation}

\begin{table}
\caption{Ablation study for the proposed regularization with L1 reward alignment. Training dataset is 40K samples from Unified-Feedback and base model is Gemma-2B-it.}
\centering
\resizebox{0.8\linewidth}{!}{
\begin{tabular}{c|c c c c c}
\toprule[1.5pt]
Loss Type  &  Chat &  Chat-Hard & Safety & Reasoning & Average \\
\midrule[1pt]
L1 loss & 96.4 & 42.1 & 80.3 & 74.3 & 73.3 \\
Ours & 95.2 & 46.9 & 83.5 & 75.2 & 75.2 \\
\bottomrule[1.5pt]   
\end{tabular}}
\label{table: L1_evaluation}
\end{table}

In this work, to ensure meaningful reward consistency between adjacent processes, we avoid using the L1 loss (which measures absolute reward differences and directly aligns rewards) due to its limitations. To demonstrate the superiority of our proposed reward consistency regularization over L1, we present comparative results in Table~\ref{table: L1_evaluation}. The experimental results show that our method achieves higher average accuracy, confirming its effectiveness.

\section{Broader impacts}
\label{sec: broader impacts}
Reward models serve critical functions in both RLHF pipelines and inference-time verification systems, playing a pivotal role in enhancing the ability to generate safer, higher-quality, and more factually accurate responses of LLMs. The focus of our work on improving reward models' generalization capabilities for unseen responses consequently makes better use of the reward model in RLHF and inference-time verification, offering significant positive societal impacts.
While comprehensive analysis reveals no immediate negative societal impacts inherent to our methodology, we acknowledge one potential secondary risk: the theoretical possibility that our generalization improvements could be repurposed by bad actors to train more harmful language models. On balance, our approach itself introduces no direct negative impacts, and the societal benefit remains positive.


\end{document}